\documentclass[letterpaper, 10 pt, conference]{ieeeconf}  %

\IEEEoverridecommandlockouts                              %

\usepackage{color}
\usepackage{times}
\usepackage{epsfig}
\usepackage{graphicx}
\usepackage{amsmath}
\usepackage{amssymb}
\usepackage[T1]{fontenc}
\usepackage{bbding}

\usepackage[utf8]{inputenc}
\usepackage{pgfplots}
\pgfplotsset{compat=newest}
\usepgfplotslibrary{groupplots}
\usepgfplotslibrary{dateplot}
\DeclareUnicodeCharacter{2212}{-}

\title{\LARGE \bf
Track to Reconstruct and Reconstruct to Track
}

\author{Jonathon Luiten$^{*1}$, Tobias Fischer$^{*1}$ and Bastian Leibe$^{1}$
\thanks{*Equal Contribution}%
\thanks{$^{1}$RWTH Aachen University \newline
{\tt \{luiten, leibe\}@vision.rwth-aachen.de \newline
tobias.fischer@rwth-aachen.de}
}%
\thanks{Code available: {\tt https://github.com/tobiasfshr/MOTSFusion}}%
}

\newcommand{\PAR}[1]{\vskip1pt \noindent {\bf #1~}}

\let\vec=\mathbf
\let\mat=\mathbf

\DeclareMathOperator{\p}{\vec{p}} %
\DeclareMathOperator{\q}{\vec{q}} %
\DeclareMathOperator{\kcov}{\mat{\Sigma}} %

\begin{document}

\maketitle
\thispagestyle{empty}
\pagestyle{empty}

\setlength{\parindent}{0pt}
\setlength\abovedisplayskip{2pt}
\setlength\belowdisplayskip{2pt}
\setlength\abovedisplayshortskip{2pt}
\setlength\belowdisplayshortskip{2pt}

\begin{abstract}
Object tracking and 3D reconstruction are often performed together, with tracking used as input for reconstruction. However, the obtained reconstructions also provide useful information for improving tracking. We propose a novel method that closes this loop, first tracking to reconstruct, and then reconstructing to track. Our approach, MOTSFusion (Multi-Object Tracking, Segmentation and dynamic object Fusion), exploits the 3D motion extracted from dynamic object reconstructions to track objects through long periods of complete occlusion and to recover missing detections.
Our approach first builds up short tracklets using 2D optical flow, and then fuses these into dynamic 3D object reconstructions. The precise 3D object motion of these reconstructions is used to merge tracklets through occlusion into long-term tracks, and to locate objects when detections are missing. 
On KITTI, our reconstruction-based tracking reduces the number of ID switches of the initial tracklets by more than 50\%, and outperforms all previous approaches for both bounding box and segmentation tracking. 
\end{abstract}

\section{INTRODUCTION}

Multi-Object Tracking (MOT) is the task of localizing objects in a video and assigning consistent IDs so that each instance of the same object is always given the same ID. This task is crucial for applications such as autonomous vehicles and mobile robots, which need to understand the presence, location and motion of dynamic objects.
MOT methods need to track objects both in frames where the objects are contiguously present, as well as through long periods where the objects are not visible due to occlusion. Many current MOT approaches focus on the first part, successfully tracking objects when they are consistently visible, but fail to track objects long-term through disappearance and occlusion, assigning incorrect IDs to objects when they reappear. However, accurate long-term tracking is crucial for understanding complex scenes to the levels required for fully autonomous systems.

\begin{figure}
	\centering
		\includegraphics[width=1.0\linewidth]{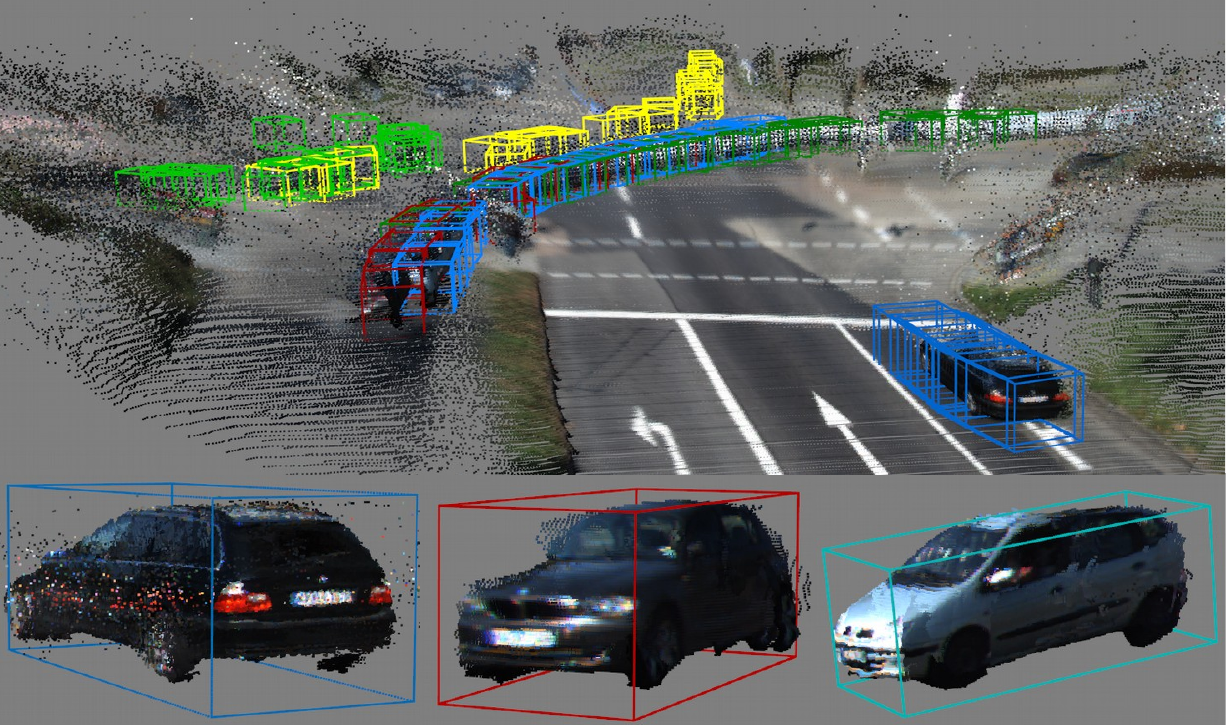}
	\caption{\textbf{Qualitative results of our dynamic 3D reconstruction based tracking.} Top: Global scene reconstruction with the 3D bounding boxes for each object track in world coordinates. Bottom: Result of our dynamic object reconstruction, for three of the object tracks in the top image.
	}
	\label{fig:1stpage}
\vspace{-19pt}
\end{figure}

\begin{figure*}[t!]
	\centering
		\includegraphics[width=1.0\textwidth]{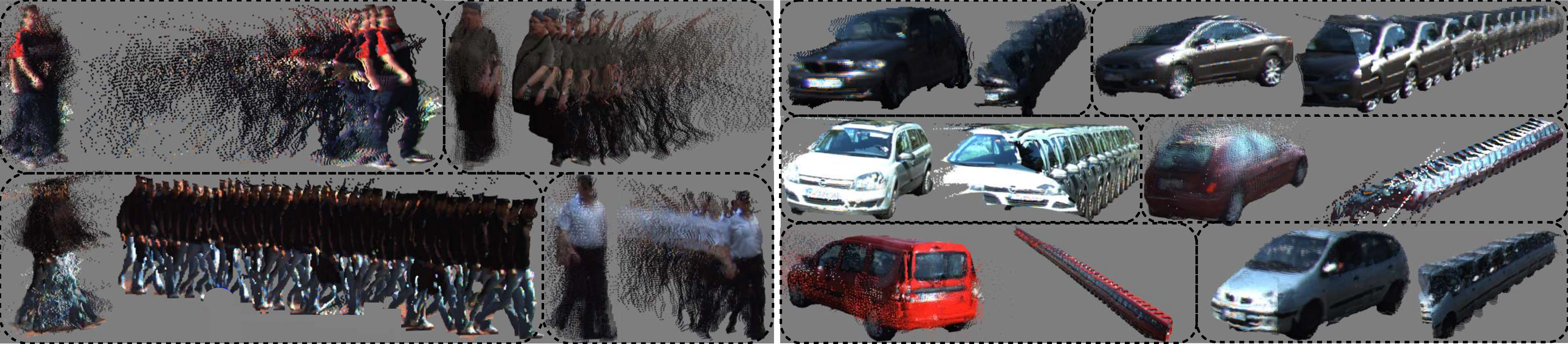}
	\caption{\textbf{Qualitative results of our dynamic 3D reconstructions which are used for tracking.} For each object, the right visualization shows every 3D point in a given tracklet in world-space. The left visualization shows the 3D reconstruction obtained when mapping each of these points into object-centric space defined by our homogeneous motion transformations. Every point in our set of tracklet masks is present in both visualizations.}
	\label{fig:merging}
\vspace{-10pt}
\end{figure*}	

To tackle long-term tracking, we propose to use dynamic 3D reconstructions to estimate the 3D motion of objects. Using this motion information, it is possible to track objects through occlusion and to recover missing detections, greatly improving long-term tracking results. Our algorithm, MOTSFusion (Multi-Object Tracking, Segmentation and dynamic object Fusion), closes the tracking-reconstruction loop, building reconstructions from tracks, and then improving tracking using these reconstructions. An example of our reconstruction-based tracking can be seen in Fig. \ref{fig:1stpage}.

MOTSFusion consists of a two-stage pipeline. First, we associate detections into short spatio-temporally consistent tracklets in the image domain by calculating a segmentation mask for each detection and measuring the consistency of these masks under a warp defined by optical flow. In a second stage, we project these tracklets into a global 3D domain using the camera ego-motion and a per pixel depth estimate. For each tracklet, a set of homogeneous transformations is calculated which align the object representations at each timestep into a dynamic reconstruction, defining the precise 3D motion of the object. Visual results of this are shown in Fig. \ref{fig:merging}. We extrapolate the 3D trajectories of each tracklet and merge tracklets into long-term object tracks by measuring the consistency of the estimated 3D trajectories. This is able to bridge long periods of occlusion and missing detections. Finally, we fill in missing detections and segmentation masks using the estimated trajectories. Our tracker exploits both 2D image space motion consistency (using optical flow and segmentation masks), and 3D world-space motion consistency (using dynamic 3D reconstructions from depth and ego-motion estimates) for accurate long-term tracking.

In order to fuse dynamic objects in 3D, MOTSFusion relies on the assumption that the objects only move with rigid body transformations. This assumption is valid for cars, but not for pedestrians which can move in an articulated manner. Despite this, our 3D tracklet merging algorithm based on the calculation of these rigid-body transformations performs very well for pedestrians. As visualized in Fig. \ref{fig:merging}, although our method cannot capture the fine-grained details of the articulated motion, it can still accurately capture the overall object motion which is what is required for tracking.

Our long-term tracking is able to reduce the number of ID switches by around 50\% compared to the initial 2D tracking results. Furthermore, our method has between 60\% and 70\% less ID switches than the previous best competing methods (\cite{beyondpixels} and \cite{mots}), using the same set of detections. This shows that our method successfully creates long-term tracks that are consistent even through occlusion, which is crucial for many applications such as autonomous vehicles.
We present several versions of MOTSFusion showing that it can be adapted to work online or offline, and use LiDAR, stereo or monocular depth input.

We make the following contributions. 
(i) We present a novel long-term tracking pipeline that uses dynamic 3D object reconstructions to merge tracks through long periods of occlusion.
(ii) We present a method for recovering missed detections of objects based on their 3D motion obtained from 3D reconstruction.
(iii) We present a thorough experimental evaluation validating the effectiveness of our reconstruction-based tracking.

\begin{figure*}[t!]
	\centering
		\includegraphics[width=1.0\textwidth]{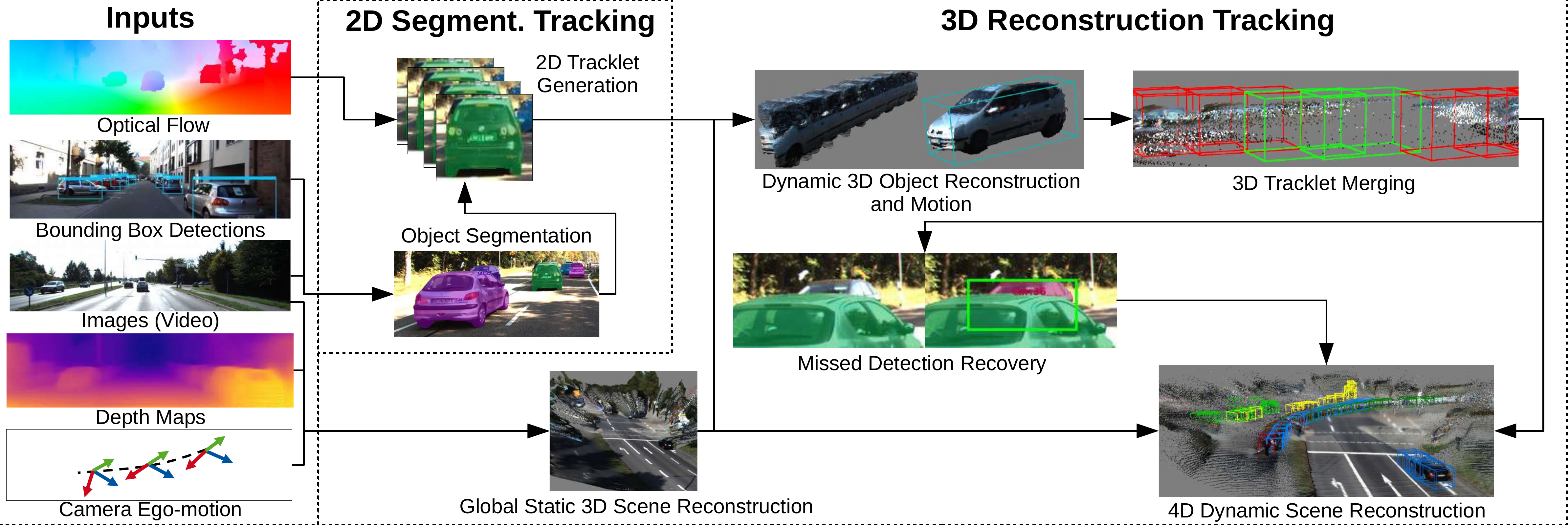}
	\caption{\textbf{Overview of the proposed MOTSFusion method.} Given a set of inputs typical for autonomous driving applications, our method performs tracking in a two-stage pipeline. First tracklets are formed using 2D image space motion consistency from optical flow and segmentation masks. Secondly 3D world-space motion consistency is used to merge tracklets together into accurate long-term tracks while recovering missed detections. This is performed by calculating the precise 3D transformations that result in a dynamic 3D object reconstruction for each object.
	}
	\label{fig:figure}
\vspace{-10pt}
\end{figure*}

\section{Related Work}
\PAR{Tracking-by-Detection.}
Multi-Object Tracking (MOT) has been tackled by many approaches.
This work builds upon many successful methods which follow a tracking-by-detection paradigm \cite{milan2013continuous, tian2018detection, chang2008, tang2017multiple, andriluka2008people}. Such methods can be distinguished by how data-association is performed, and which features are used for association. Data-association for MOT has been approached by network flow \cite{Zhang08CVPR}, multi-hypothesis tracking \cite{Kim15ICCV}, quadratic pseudo boolean optimization \cite{leibe2008} and conditional random fields \cite{NOMT, CIWT}. Previous methods have exploited the use of 2D motion consistency \cite{NOMT}, 3D motion consistency \cite{CIWT}, and visual embedding similarity \cite{premvos} as features for performing association.

\PAR{Hierarchical 2D/3D Tracking.}
For data-association, we adopt a hierarchical tracklet creation and merging approach, similar to \cite{chang2008,kaucic2005} and \cite{geiger20143d}, who have shown that such a multi-stage approach can successfully perform tracking.
In contrast to these previous approaches, we separate which features are used for association at each stage. First we use only 2D motion consistency for tracklet creation, and then only 3D motion consistency for tracklet merging. Previous approaches \cite{geiger20143d} have used appearance models of visual similarity for long-term tracking. We rely solely on 2D and 3D motion consistency, showing that motion cues are enough to perform long-term tracking without exploiting visual similarity.

\PAR{Segmentation Tracking.}
Recently segmentation masks have been exploited for improving tracking and creating pixel-accurate tracking results \cite{mots,CAMOT,premvos}. These often use optical flow \cite{dispflownet3} or scene flow \cite{wedel2011stereoscopic} to model the motion of each pixel. We adopt both of these techniques using optical flow and scene flow within an object mask to determine motion consistency in both 2D and 3D. We go beyond these methods and use the per-pixel depth values within an the object mask to create 3D object reconstructions over time.

\PAR{Tracking in 3D.}
Producing 3D tracking results is invaluable for robotics and autonomous vehicles. There has been a large body of research \cite{CIWT,CAMOT,beyondpixels,held2013precision, luo2018fast,mitzel2012taking} in this area.
Most methods \cite{CIWT, beyondpixels} track using 3D bounding boxes and rely on simple motion cues such as average scene flow vectors. A few methods \cite{held2013precision,mitzel2012taking} have tried to track object reconstructions in 3D by determining the precise transformation for an object between frames. Our method follows a similar idea, however unlike previous methods that used techniques like ICP \cite{besl1992icp} for aligning point-clouds, our approach is able to directly optimize for an alignment of points whose correspondences are given by optical flow.	
	
\PAR{Dynamic Object Reconstruction.}
Dynamic object reconstruction is closely related to MOT, as to perform reconstruction first tracking needs to be performed. A number of methods \cite{Barsan2018ICRA,maskfusion} have approached this task. Compared to these methods, our method closes the loop; using these reconstructions to further improve tracking.	

\section{Our Approach}
MOTSFusion is built upon four key ideas. 
(i) Accurate short tracklets can be constructed using the 2D motion consistency obtained from segmentation masks and optical flow \cite{premvos,mots}. This results in highly precise short tracklets for when objects are contiguously visible. 
(ii) For an object undergoing rigid-body transformations, there exists a set of such transformations that fuses all of the point clouds of this object from each timestep into a consistent 3D reconstruction, and these transformations define the 3D motion of that object.
(iii) Even for non-rigid objects, if the best-fit rigid-body transformation is calculated in a robust way, this results in 3D motion estimates which are accurate enough to determine the overall 3D motion of an object. 
(iv) The consistency between the estimated 3D motion of two tracklets contains enough information to determine if these tracklets belong to the same object and should be merged into one longer track.

\begin{figure*}[t!]
	\centering
		\includegraphics[width=1.0\textwidth]{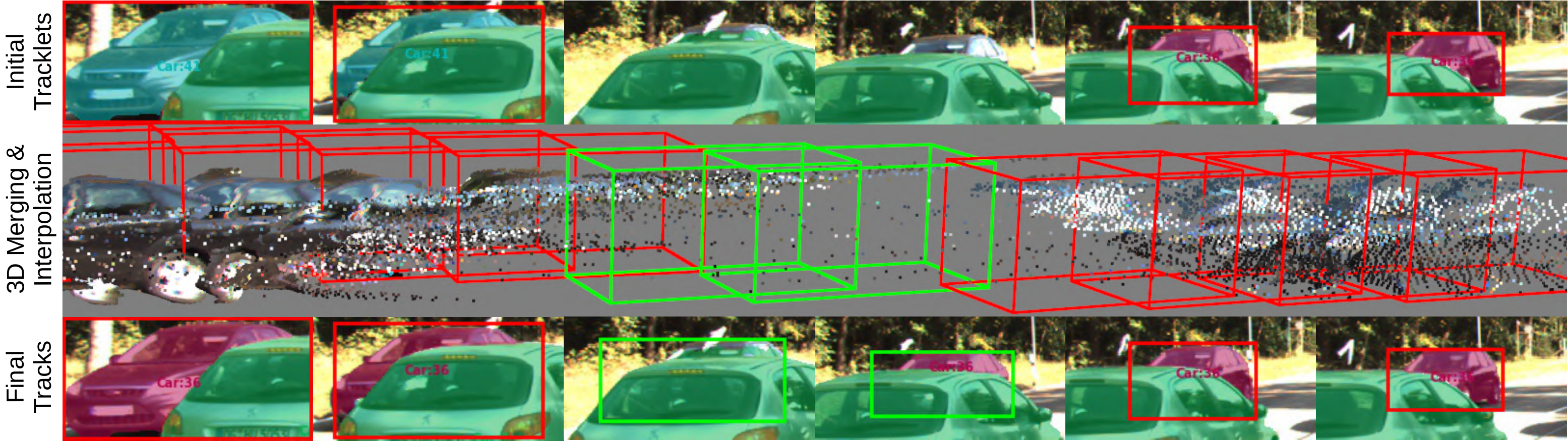}
	\caption{\textbf{Example of our 3D tracklet merging and missing detection filling} Top: Initial 2D bounding box detections (red) and segmentation based tracklet results. Middle: Result of our interpolated 3D bounding boxes between the two tracklets, showing the temporal consistency of the two tracklets' 3D motion through the frames with no detections. Bottom: Original 2D bounding box detections (red) and the new interpolated 2D bounding boxes (green) together with the filled segmentation mask and merged track ID.
	}
	\label{fig:teaser}
\vspace{-15pt}
\end{figure*}

Following these ideas, we develop MOTSFusion (shown in Fig. \ref{fig:figure}), a two-stage algorithm which first creates short tracklets using the 2D motion consistency of segmentation masks under an optical flow warp, and then fuses these tracklets, using depth and ego-motion estimates, into consistent dynamic 3D object reconstructions. The transformations required for these reconstructions are then used to estimate the 3D motion of object tracklets, merging them into longer tracks if they undergo consistent 3D motion, and using the extrapolated positions to fill in detections where they are missing.
MOTSFusion simultaneously tracks objects as both image space segmentation masks and world-space 3D object reconstructions. An example of 3D merging and interpolation is shown in Fig. \ref{fig:teaser}. A number of 3D object reconstruction results are shown in Fig. \ref{fig:merging}.

\PAR{Tracking Inputs.}
MOTSFusion uses video frames as input, as well as per frame ego-motion and depth estimates. For our main experiments, we use stereo depth (DispNet3 \cite{dispflownet3}), and ego-motion from a SLAM algorithm (ORB-SLAM2 \cite{orbslam2}). However, MOTSFusion is designed to work with any available depth estimates (stereo, LiDAR, RADAR, SfM or monocular) and ego-motion estimates (SLAM, GPS, IMU). We also show results when using LiDAR and single-image depth estimates.
For object detection, we use both the recurrent rolling convolution (RRC) detector \cite{rrc_detector} and Track R-CNN \cite{mots}.
We use optical flow obtained from \cite{dispflownet3}.

\PAR{Bounding Boxes to Segmentation Masks.}
We initially estimate a segmentation mask for each bounding box detection. We use a fully convolutional neural network from \cite{premvos} which we call BB2SegNet. This crops and resizes an image region given by a bounding box to a $385\times385$ patch and outputs a segmentation mask for each box.

\PAR{2D Tracklet Generation.}
We generate tracklets by warping the pixels of each segmentation mask using the optical flow values at these pixels into the next frame. We calculate the IoU (intersection over union) of these warped masks and the set of segmentation masks in the next frame to create association similarities. We use the Hungarian algorithm to assign masks to previously existing tracklets. All masks that are not merged into previous tracklets begin new tracklets. A minimum IoU threshold is required to merge masks.

\PAR{4D Scene Reconstruction.}
In order to obtain 3D object motion, we need the 3D location of each object in each timestep in a common world-frame. For our world-frame we use the position of the camera in the first frame and create a 4D (3D + time) point cloud of the scene using depth $d_t$, camera intrinsics $K \in \mathbb{R}^{3\times 3}$ and camera position matrix $T_t \in \mathbb{R}^{4\times 4}$ (the homogeneous transformation matrix of the accumulated ego-camera position over time) with the following equations:
\begin{equation}
(x,y,z)^C = ((u,v,1)\times(K^{-1})^\top) \cdot d_t 
\label{eq:project}
\end{equation}
\begin{equation}
(x,y,z,1)^W =  (x,y,z,1)^C \times T_t^\top
\label{eq:world}
\end{equation}
where $(u,v)$ is the pixel location at time $t$, $(x,y,z)^C$ is the 3D point corresponding to pixel $(u,v)$ in the current camera frame and the $(x,y,z)^W$ is the 3D point in the world frame. This 4D scene reconstruction is used for estimating the motion of objects independent of ego-camera motion, and for visualizing 3D tracking (Fig. \ref{fig:1stpage}).

\PAR{Dynamic 3D Object Fusion.}
We calculate a set of rigid-body transformations that warp the set of 3D points for an object tracklet in each time-step into a consistent 3D object reconstruction.
We fit a homogeneous transformation from the world-space points within a pixel-mask from time $t$ to the world-space points within the pixel-mask of the same tracklet at time $t+1$. This transformation is accumulated over all timesteps so that all points in an object tracklet are warped into a consistent 3D reconstruction.

To minimize the influence of incorrect depth estimates and object masks, we filter the set of 3D points used to fit the transformation by calculating the local outlier factor $LOF_k$ \cite{localoutfac} using the local reachability density $\mathit{lrd}_k$ of each point $\p$ with its $k$ nearest neighbors $N_k$. The local reachability density uses the reachability distance $\mathit{rd}_k$, which is defined by the Minkowski distance $d_{min}$ between two points $\p$ and $\q$ and the $k$-distance $d_k$ of $\q$, which is the maximum of the Minkowski distances of the $k$ nearest neighbors of $\q$, as follows:
\begin{equation}
\mathit{rd}_k(\p, \q) = \max\{d_k(\q), d_{min}(\p, \q)\}
\end{equation}
\begin{equation}
\mathit{lrd}_k(\p) := 1 / (\frac{{\sum}_{\q \in N_k} \mathit{rd}_k(\p, \q)}{|N_k(\p)|}) 
\end{equation}
Using the local reachability distance of each point, we can compute the local outlier factor $\mathit{LOF}_k$ \cite{localoutfac} by:
\begin{equation}
\mathit{LOF}_k(\p) := \frac{{\sum}_{\q \in N_k} \frac{\mathit{lrd}_k(\q)}{\mathit{lrd}_k(\p)}}{|N_k(\p)|}
\end{equation}
Subsequently, we filter out points with a higher $\mathit{LOF}_k$ than the median $\mathit{LOF}_k$ of all points.
For the remaining 3D points at time $t$ and $t+1$, we use the optical flow vector to correlate points between the two timesteps. We further filter these points to only those for which correspondences exist between the two timesteps. We then sample a maximum of $200$ corresponding points over the mask.

On these points we perform a non-linear least-squares optimization by minimizing the $L2$ distance between each pair of corresponding points in the two point clouds to determine the homogeneous transformation that best aligns the two 3D point clouds. We restrict this transformation to be a 3-DoF homogeneous transformation (X and Z translation, and rotation around the ground plane normal). This simplification is valid for KITTI as the ground-plane is approximately flat. We parametrize the 3D rigid motion with $\xi \in \mathfrak{se}(2)$, the Lie algebra associated with $SE(2)$, which is a minimal representation for this motion.
This transformation gives the precise alignment (motion) of an object between timesteps in world coordinates. For each tracklet, we calculate this transformation for all pairs of neighboring frames and accumulate the transformations, merging the point clouds from every timestep into one 3D reconstruction in a consistent object-centric reference frame. Results of this are shown in Fig. \ref{fig:merging}, where it is evident that our transformations are accurate, even when accumulated over many timesteps, and even for objects undergoing non-rigid transformations.

\PAR{3D Tracklet Merging.}
Tracklets are merged by analyzing their 3D consistency under the motion given by the object fusion.
We examine merging candidates for a terminated tracklet up to $N$ frames beyond the terminated tracklet's temporal extent. The inherent uncertainty of the motion estimates is measured by the alignment accuracy of the rigid-body transformations using the residual error of the non-linear fitting. We define a `Trusted Motion Region' (TMR) as the set of contiguous motion transformations closest to the end of a tracklet which are all below a residual error threshold (requiring at least two transformations contiguously). We map each of the transformations $\xi = (\xi_x, \xi_z, \xi_\theta)$ in the TMR from global coordinates into object-centric coordinates $\hat{\xi}$ which are centered at the current object center $\p$ as follows:
\begin{equation}
\begin{pmatrix} \hat{\xi}_x \\ \hat{\xi}_z \\ \hat{\xi}_\theta \end{pmatrix} = 
\begin{pmatrix} \cos(\xi_\theta)-1 & -\sin(\xi_\theta) & \xi_x \\ \sin(\xi_\theta) & \cos(\xi_\theta)-1 & \xi_z \\ 0 & 0 & \xi_\theta \end{pmatrix} 
\begin{pmatrix} \p_x \\ \p_z \\ 1\end{pmatrix}
\end{equation}
This object-centric motion parametrization $\hat{\xi}$ encodes the same transformation as the original $\xi$, but now is centered on the moving object center. This allows us to meaningfully average these transformations, as well as to extrapolate them into the future, always relative to the current object center estimate. It is not possible to meaningfully average or extrapolate homogeneous transformations in global coordinates, they must first be transformed into an object-centric coordinate system.

We use the median location of our filtered 3D point set as our object center $\p$. We also model the uncertainty of the object position from stereo estimation.  
We model an object's location at each timestep using a multivariate normal distribution. The mean is $\p$ and the covariance, $\kcov$, is obtained using the following \cite{multi_view_geo}:
\begin{equation}
\kcov = \begin{pmatrix} \mathbf{F}_{L}^{\top}(\p) \kcov_{\text{pix}}^{-1} \mathbf{F}_{L}(\p) + \mathbf{F}_{R}^{\top}(\p) \kcov_{\text{pix}}^{-1} \mathbf{F}_{R}(\p) \end{pmatrix}^{-1},
\label{eq:uncertainty}
\end{equation}
where $\mat{F}_{L}(\p), \mat{F}_{R}(\p) \in \mathbb{R}^{2\times 3}$ are Jacobians (evaluated at $\p$) of the left and right camera projection matrices, respectively.
The matrix $\kcov_{\text{pix}} = \begin{pmatrix} \sigma_u\; 0; 0\; \sigma_v \end{pmatrix}$  models uncertainty in the pixel measurements, with $\sigma_u$ and $\sigma_v$ both set to 0.5.

The 3D motion consistency between two tracklets is calculated by extrapolating both tracklets' 3D motion towards each other using the average relative transformation of the TMRs into all timesteps between the TMRs of the two tracklets, including the last frame of the earlier TMR and the first frame of the latter. From these 3D position estimates and their uncertainties, the 3D motion consistency of two tracklets is given by the average Mahalanobis distance of these 3D position estimates weighted by their respective uncertainties from the covariances $\kcov$ (Eq. \ref{eq:uncertainty}).
If the extrapolated motion for one tracklet cannot be estimated robustly due to lack of a TMR, we determine the 3D motion consistency by only extrapolating one towards the other and computing the consistency in the timestep of the last frame of the tracklet for which we have no motion estimate.
If both tracklets do not have a robust motion estimate, then both are assumed to be stationary.
We merge two tracklets if the 3D motion consistency between them is over a threshold.

\PAR{Missing Detection Recovery.}
For each frame between two merged tracklets, we wish to determine if the object is visible and the detection was missed, or if it is actually completely occluded.
We first estimate a 3D bounding box for each frame. We use $\p$ as the bounding box center and assume fixed dimensions for pedestrians and cars given by the average width/height/length of the 3D bounding boxes in the KITTI 3D detection training set. This simplifies 3D bounding box estimation, and results in adequate localization for producing segmentations. We set the bounding box orientation $\theta$ to the direction of motion if the object undergoes significant motion. Otherwise, we set $\theta$ to the direction of the eigenvector with the largest eigenvalue in bird's eye view for all of the 3D points in the object in the current timestep (i.e. the direction of the greatest variance). Examples of 3D bounding boxes can be seen in Fig. \ref{fig:1stpage}.

We project this 3D bounding box into image space as a 2D bounding box and average it with the previous-frame bounding box corners warped by the median optical flow vector of all points in the bounding box. We run BB2SegNet to get a segmentation mask for this box. We check the validity of this segmentation mask by taking the depth values for the points within the new mask and projecting these back into 3D world coordinates using Eq. \ref{eq:project} and Eq. \ref{eq:world}. If the 3D points of the new mask are sufficiently close to the 3D bounding box center, then the mask passes the consistency check. This check determines if the object has been occluded, since if occluded the points belonging to the new mask will be significantly in front of the estimated 3D bounding box. This is used to fill in missing detections, without introducing many false positives. As well as running missing detection recovery between merged tracklets, we also run it at the beginning and end of whole tracks where we apply this procedure to every frame until we reach a frame where the consistency check fails, the object moves out of the camera field of view, or we reach the end of the video sequence.

\PAR{Online Version of MOTSFusion.}
MOTSFusion extrapolates the trajectories of objects both forward and backward in time to perform long-term tracking. Thus, by default this method works in an offline setting. However, such a reconstruction-based tracking approach can be adapted to work online for use in applications such as autonomous driving and robotics. In order to create an online version of MOTSFusion we simply only extrapolate the trajectories of objects forward in time instead of both forward and backwards. At each new timestep we then match the current detections with the forward extrapolated 3D trajectories, instead of matching the trajectories of two tracklets being extrapolated forward and backward. Finally, for missing detection recovery we also only fill in missing detections in the current frame based on the forward motion estimate. The initial 2D tracklet building step is always performed online.

\begin{table*}
\resizebox{\linewidth}{!}{
\centering
\small
\begin{tabular}{|c|c|c|c||c|c|c|c|c||c|c|c|c|c|}
\cline{5-14} 
\multicolumn{4}{c|}{} & \multicolumn{5}{c||}{Cars} & \multicolumn{5}{c|}{Pedestrians}\tabularnewline
\hline 
Tracking & Detect. & Segm. & Speed & sMOTSA & MOTSA & IDS & FP & FN & sMOTSA & MOTSA & IDS & FP & FN\tabularnewline
\hline 
\hline

\textbf{Ours} & RRC\cite{rrc_detector} & BB2SegNet\cite{premvos} & 0.44 & \textbf{85.7} & \textbf{94.5} & \textbf{31} & 44 & 364 & - & - & - & - & -\tabularnewline
\hline 
Ours (no fill) & RRC\cite{rrc_detector} & BB2SegNet\cite{premvos} & 0.43 & 85.6 & 94.3 & \textbf{31} & \textbf{37} & 386 & - & - & - & - & -\tabularnewline
\hline 
Ours (2D) & RRC\cite{rrc_detector} & BB2SegNet\cite{premvos} & \textbf{0.14} & 85.2 & 94.0 & 61 & \textbf{37} & 386 & - & - & - & - & -\tabularnewline
\hline 
Ours (online) & RRC\cite{rrc_detector} & BB2SegNet\cite{premvos} & 0.44 & 85.5 & 94.3 & 35 & 38 & 385 & - & - & - & - & -\tabularnewline
\hline
BePix\cite{beyondpixels} & RRC\cite{rrc_detector} & BB2SegNet\cite{premvos} & 0.36 & 84.9 & 93.8 & 97 & 61 & \textbf{337} & - & - & - & - & -\tabularnewline
\hline 
Oracle & RRC\cite{rrc_detector} & BB2SegNet\cite{premvos} & - & \textit{86.9} & \textit{95.9} & \textit{0} & \textit{3} & \textit{330} & - & - & - & - & - \tabularnewline
\hline 
\hline

\textbf{Ours} & TrRCNN\cite{mots} & BB2SegNet\cite{premvos} & 0.44 & \textbf{82.8} & \textbf{90.5} & \textbf{51} & 51 & \textbf{661} & \textbf{59.4} & \textbf{72.6} & \textbf{35} & 99 & \textbf{784}\tabularnewline
\hline 
Ours (no fill) & TrRCNN\cite{mots} & BB2SegNet\cite{premvos} & 0.43 & 82.6 & 90.2 & \textbf{51} & \textbf{41} & 695 & 58.8 & 71.8 & \textbf{35} & \textbf{94} & 814 \tabularnewline
\hline 
Ours (2D) & TrRCNN\cite{mots} & BB2SegNet\cite{premvos} & \textbf{0.14} & 81.9 & 89.6 & 102 & \textbf{41} & 695 & 58.2 & 71.2 & 55 & \textbf{94} & 814 \tabularnewline
\hline
Ours (online) & TrRCNN\cite{mots} & BB2SegNet\cite{premvos} & 0.44 & 82.6 & 90.2 & \textbf{51} & 45 & 688 & 58.9 & 71.9 & 36 & 95 & 810 \tabularnewline
\hline 
Oracle & TrRCNN\cite{mots} & BB2SegNet\cite{premvos} & - & \textit{87.0} & \textit{96.0} & \textit{0} & \textit{19} & \textit{303} & \textit{68.7} & \textit{86.0} & \textit{0} & \textit{29} & \textit{440} \tabularnewline
\hline
\hline

\textbf{Ours} & TrRCNN\cite{mots} & TrRCNN\cite{mots} & 0.44 & \textbf{78.2} & \textbf{90.0} & \textbf{36} & 94 & 673 & \textbf{50.1} & \textbf{68.0} & \textbf{34} & 181 & 855\tabularnewline
\hline 
Ours (no fill) & TrRCNN\cite{mots} & TrRCNN\cite{mots} & 0.43 & 78.1 & 89.8 & \textbf{36} & \textbf{86} & 699 & 49.5 & 67.2 & \textbf{34} & \textbf{178} & 886\tabularnewline
\hline 
Ours (2D) & TrRCNN\cite{mots} & TrRCNN\cite{mots} & \textbf{0.14} &  77.5 & 89.2 & 85 & \textbf{86} & 699 & 48.9 & 66.6 & 53 & \textbf{178} & 886\tabularnewline
\hline
Ours (online) & TrRCNN\cite{mots} & TrRCNN\cite{mots} & 0.44 & 78.1 & 89.8 & 44 & 87 & 686 & 49.5 & 67.3 & 35 & \textbf{178} & 882\tabularnewline
\hline  
BePix\cite{beyondpixels} & RRC\cite{rrc_detector} & TrRCNN\cite{mots} & 0.36 & 76.9 & 89.7 & 88 & 280 & \textbf{458} & - & - & - & - & -\tabularnewline
\hline
TrRCNN\cite{mots} & TrRCNN\cite{mots} & TrRCNN\cite{mots} & 0.50 & 76.2 & 87.8 & 93 & 134 & 753 & 46.8 & 65.1 & 78 & 267 & \textbf{822}\tabularnewline
\hline 
CIWT\cite{CIWT} & TrRCNN\cite{mots} & TrRCNN\cite{mots} & 0.28 & 68.1 & 79.4 & 106 & 333 & 1214 & 42.9 & 61.0 & 42 & 401 & 863\tabularnewline
\hline 
CAMOT\cite{CAMOT} & TrRCNN\cite{mots} & TrRCNN\cite{mots} & 0.76 & 67.4 & 78.6 & 220 & 172 & 1327 & 39.5 & 57.6 & 131 & 198 & 1090\tabularnewline
\hline 
Oracle & TrRCNN\cite{mots} & TrRCNN\cite{mots} & - & \textit{82.0} & \textit{95.1} & \textit{0} & \textit{30} & \textit{361} & \textit{58.3} & \textit{80.3} & \textit{0} & \textit{23} & \textit{635}\tabularnewline
\hline

\end{tabular}
}
\caption{\textbf{Mask tracking results on KITTI MOTS Validation.} We do not evaluate the RRC detector on pedestrians, because detections are only available for cars. Best result numbers per section in \textbf{bold}. Speed is measured in seconds per frame.
	}
\label{fig:table_comp_mask}
\vspace{-25pt}
\end{table*}

\begin{table}
\centering
\setlength{\tabcolsep}{1.2pt}
\resizebox{1.0\linewidth}{!}{
\begin{tabular}{|c||c|c|c|c|c||c|c|c|c|c|}
\cline{2-11} 
\multicolumn{1}{c|}{} & \multicolumn{5}{c||}{Cars} & \multicolumn{5}{c|}{Pedestrians}\tabularnewline
\hline 
Tracking & sMOTSA & MOTSA & IDS & FP & FN & sMOTSA & MOTSA & IDS & FP & FN\tabularnewline
\hline 
\hline 

\textbf{Ours} & \textbf{75.0} & \textbf{84.1} & \textbf{201} & \textbf{295} & \textbf{5342} & \textbf{58.7} & \textbf{72.9} & \textbf{279} & \textbf{465} & \textbf{4870}\tabularnewline
\hline 
TrRCNN\cite{mots} & 67.0 & 79.6 & 692 & 1310 & 5479 & 47.3 & 66.1 & 481 & 1179 & 5363\tabularnewline
\hline 
\end{tabular}
}
\caption{\textbf{Mask tracking results of our best method on KITTI MOTS Test.}
}
\label{fig:table_comp_test-mask}
\vspace{-25pt}
\end{table}

\section{Experiments}
\label{sec:exp}
\PAR{Datasets.}
We evaluate MOTSFusion using the KITTI dataset \cite{kitti_mot}, which contains traffic scenes captured from a moving vehicle. We use the KITTI tracking benchmark to evaluate our tracking results for both cars and pedestrians in real-world driving scenes. 
The annotations for this have been extended with pixel-level mask annotations to evaluate the MOTS task (multi-object tracking and segmentation) \cite{mots}.
We use the official KITTI test server as well as the validation split from \cite{mots} for evaluation.

\PAR{Evaluation Metrics.}
We adopt the CLEARMOT \cite{clearmot} metrics which are the standard for KITTI MOT \cite{kitti_mot}. For bounding box tracking methods are ranked by MOTA, which incorporates three error types: false positives (FP), false negatives (FN) and ID switches (IDS).
For segmentation tracking we adopt a version of these adapted for masks \cite{mots}, where methods are ranked by MOTSA, the segmentation version of MOTA, as well as sMOTSA, which takes into account the segmentation accuracy by incorporating the IoU of the predicted and ground truth masks into the score.

The definition of ID switches is not consistent between benchmarks. For segmentations we use the IDS version from \cite{mots} (also used in MOTChallenge \cite{milan2014mot}). For bounding boxes we present the original version from KITTI \cite{kitti_mot} as IDS (and MOTA respectively), however it has often been pointed out that this definition does not correctly account for ID switches \cite{bento2016metric, shitrit2011tracking, Yu_2016_CVPR, maksai2017non}. Hence, we also present results on the validation set using the definition from \cite{mots} and \cite{milan2014mot}, which we label IDS* and MOTA*, respectively.
This definition counts ID switches not only when the ID label switches between two contiguous frames, but also when there is a gap (e.g. occlusion) in between an ID switch occurring.

\PAR{Segmentation Tracking: Setup comparison.}
In Table \ref{fig:table_comp_mask} we present results of MOTSFusion for segmentation tracking on the KITTI MOTS validation set. We compare the use of three different combinations of detectors and segmentation methods. We use both the detections and segmentations from \cite{mots} for a fair comparison to the previous state-of-the-art. We also use a stronger detector \cite{rrc_detector} and a stronger segmentation method \cite{premvos}, in order to compare to the previous state-of-the-art bounding box tracker \cite{beyondpixels} with added segmentation masks. We also report results with the detections from \cite{mots} with segmentations from \cite{premvos}. We only evaluate cars when using detections from \cite{rrc_detector} (as pedestrian detections are not available). Using better detections and better segmentations does benefit our method. Using \cite{premvos} segmentations instead of \cite{mots}, sMOTSA increases from 78.2 to 82.8 for cars and from 50.1 to 59.4 for pedestrians (4.6 and 9.3 percentage points). Using \cite{rrc_detector} detections instead of \cite{mots}, sMOTSA increases another 2.9 percentage points from 82.8 to 85.7 for cars.

\PAR{Segmentation Tracking: Method ablation.}
In all five experimental setups, we present ablation results showing how our 3D reconstruction-based tracking improves over the initial tracklets (\textit{Ours (2D)}). In the best setups, with reconstruction-based tracklet merging only (\textit{Ours (no fill)}), we reduce the IDS from 61 to 31 for cars and from 55 to 35 for pedestrians (49\% and 36\% relative improvement, respectively), while not changing the number of FPs and FNs. We observe similar results over all setups resulting in a sMOTSA increase of between 0.4 and 0.7 percentage points. This is a significant improvement in long-term tracking ability. When also performing missing detection recovery (\textit{Ours}), we reduce the number of FNs from 386 to 364 for cars and from 814 to 784 for pedestrians, while increasing the FPs only marginally from 37 to 44 for cars and from 94 to 99 for pedestrians. Again, we observe similar results over all setups resulting in a further sMOTSA increase of between 0.1 and 0.6 percentage points.

\PAR{Segmentation Tracking: State-of-the-art comparison.}
Table \ref{fig:table_comp_mask} shows that on the validation set MOTSFusion outperforms TrackR-CNN \cite{mots} with the same detections and segmentations by 2 percentage points for cars (78.2 vs. 76.2) and 3.3 for pedestrians (50.1 vs. 46.8) in sMOTSA, and has 61\% (36 vs. 93) and 56\% (34 vs. 78) less IDS for cars and pedestrians, respectively. When comparing to BeyondPixels \cite{beyondpixels} with added segmentations from BB2SegNet, we improve from 84.9 to 85.7 in sMOTSA while reducing the number of IDS from 97 to 31. This is less than a third of the IDS. Using the segmentations from \cite{mots} our method outperforms \cite{beyondpixels} by 1.3 percentage points in sMOTSA (78.2 vs. 76.9), even though \cite{beyondpixels} uses the stronger RRC detector. It also significantly outperforms other tracking methods \cite{CIWT} and \cite{CAMOT} using the same detections and segmentations. In Table \ref{fig:table_comp_test-mask} we also evaluate the best versions of MOTSFusion on the MOTS test server, where we significantly outperform the previous best results from \cite{mots}.

\PAR{Segmentation Tracking: Online version.}
In Table \ref{fig:table_comp_mask} we also present results for the online version of MOTSFusion for all five experimental setups. In terms of IDS, the online version works almost as well as the offline version, resulting in negligibly more ID switches over the five setups. This shows the strength of our reconstruction-based tracking approach, even in an online setting. However, the online version cannot leverage the same priors as the offline version for missing detection recovery, where two merged tracklets could have missing measurements in between them, resulting in similar performance to the offline version without detection filling and thus is more reliant on good detections in each frame.

\PAR{Segmentation Tracking: Oracle comparison.}
We present results for each setup using an `oracle' tracker, that takes the detections and segmentations, and associates these into tracks using the ground-truth. Our best performing method for cars obtains a sMOTSA only 1.2 percentage points below the theoretical maximum of a perfect tracker. The oracle still has FPs because the MOTS evaluation script constrains all masks in a frame to not overlap, which can cause masks that have been matched to the ground truth to become unmatched.

\PAR{Segmentation Tracking: Depth input ablation.}
MOTSFusion works with any depth input, but uses stereo \cite{dispflownet3} for the main experiments. In Table \ref{table:table_depth_ablation} we also provide results using LiDAR measurements and monocular single-image depth estimates \cite{monodepth2}. LiDAR depth is much more accurate than stereo, but it is also very sparse and may suffer from camera-LiDAR calibration issues, whereas monocular depth is usually much less accurate. To illustrate this we compute the end-point error (EPE) of the depth estimates w.r.t. to the LiDAR measurements (for pixels where this is available) and calculate the density of the LiDAR points projected into the image frame. MOTSFusion performs similarly using LiDAR as when using stereo. This shows that our method is robust to the inaccuracies present in stereo depth, and that it can successfully handle sparse input. When using the much less accurate monocular depth estimates MOTSFusion does not perform as accurately, but still improves over the 2D version which does not use any depth estimation. This shows the robustness of our method to noisy depth input.

\begin{table}
\centering
\resizebox{1.0\linewidth}{!}{
\begin{tabular}{|c|c|c||c|c|c|c|c|}
\cline{4-8} 
\multicolumn{3}{c|}{} & \multicolumn{5}{c|}{Cars}\tabularnewline
\hline 
Depth & EPE & Density & sMOTSA & MOTSA & IDS & FP & FN\tabularnewline
\hline 
\hline 
LiDAR & 0.00 & 4.1\% & 85.7 & 94.5 & 31 & 44 & 367\tabularnewline
\hline 
Stereo\cite{dispflownet3} & 1.50 & 100\% & 85.7 & 94.5 & 31 & 44 & 364\tabularnewline
\hline 
Mono\cite{monodepth2} & 2.11 & 100\% & 85.3 & 94.1 & 51 & 59 & 360\tabularnewline
\hline 
None (2D) & - & - & 85.2 & 94.0 & 61 & 37 & 386\tabularnewline
\hline 
\end{tabular}
}
\caption{\textbf{Mask tracking results on KITTI MOTS validation with different depth estimates} using RRC \cite{rrc_detector} detections and BB2SegNet\cite{premvos} segmentations. EPE is the average end-point error compared to the LiDAR depth.
	}
\label{table:table_depth_ablation}
\vspace{-20pt}
\end{table}

\begin{figure}
\centering

\includegraphics[width=1.0\linewidth]{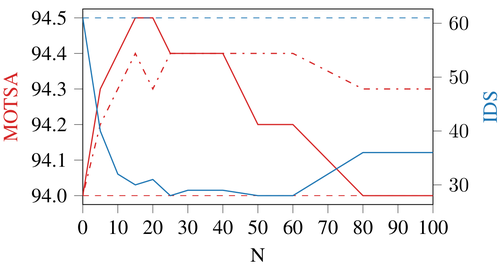}

\vspace{-2mm}
\caption{\textbf{Ablation showing the effect of varying the temporal gap size}, $N$, which is the maximum number of frames used for 3D tracklet merging. We present results for both MOTSA (red) and IDS (blue). The dashed lines show the values of our 2D tracklet generation without our reconstruction-based tracking. The dash-dotted line shows the MOTSA when performing tracklet merging but no missing detection recovery.}
\label{fig:future_ablation}
\vspace{-10pt}
\end{figure}

\PAR{Segmentation Tracking: Tracklet merging ablation.}
Fig. \ref{fig:future_ablation} shows the results of using different thresholds for the maximum number of frames $N$ between merging candidates. Compared to the initial tracklets (dashed lines), the results quickly and dramatically improve when merging tracklets up to $15$ timesteps apart for both the MOTSA score and the number of IDS. When merging beyond $25$ frames the missing detection recovery becomes less accurate, introducing more false positives, while the tracklet merging remains accurate keeping the IDS low. When merging beyond $60$ frames the tracklet merging also introduces errors and the IDS increases. For our main experiments we choose a threshold of $20$ frames. The gap between the solid red curve and the dash-dotted curve shows how much improvement over the 2D results comes from tracklet merging compared to what comes from missing detection recovery.

\begin{table}
\centering
\resizebox{1.0\linewidth}{!}{
\begin{tabular}{|c|c||c|c|c|c|c|c|}
\cline{3-8} 
\multicolumn{1}{c}{} & \multicolumn{1}{c|}{} & \multicolumn{6}{c|}{Cars}\tabularnewline
\hline 
Tracking & Speed & MOTA & MOTA{*} & IDS & IDS{*} & FP & FN\tabularnewline
\hline 
\hline 
Ours (no fill) & 0.43 & \textbf{94.0} & \textbf{93.7} & \textbf{9} & \textbf{31} & \textbf{45} & 400\tabularnewline
\hline 
Ours (2D) & \textbf{0.14} & 93.9 & 93.3 & 12 & 58 & \textbf{45} & 400\tabularnewline
\hline 
BePix\cite{beyondpixels} & 0.30 & 93.7 & 92.9 & 31 & 91 & 88 & \textbf{354}\tabularnewline
\hline 
Oracle & - & \textit{95.8} & \textit{95.8} & \textit{6} & \textit{6} & \textit{0} & \textit{311}\tabularnewline
\hline 
\end{tabular}
}
\caption{\textbf{Bounding box tracking results on KITTI Tracking Validation (Cars).} The MOTA* and IDS* metrics use the adapted IDS definitions from \cite{mots}. All methods use RRC detections \cite{rrc_detector}. Speed is measured in seconds per frame.
}
\label{fig:table_comp_val}
\vspace{-20pt}
\end{table}

\begin{table}
\centering
\begin{flushright}
\resizebox{1.0\linewidth}{!}{
\begin{tabular}{|c|c|c||c|c|c|c|}
\cline{4-7} 
\multicolumn{3}{c|}{} & \multicolumn{4}{c|}{Cars}\tabularnewline
\hline 
Tracking & Detect. & Speed & MOTA & IDS & FP & FN\tabularnewline
\hline 
\hline 
Ours & RRC\cite{rrc_detector} & 0.44 & \textbf{84.83} & \textbf{275} & \textbf{681} & 4260\tabularnewline
\hline 
BePix\cite{beyondpixels} & RRC\cite{rrc_detector} & 0.30 & 84.24 & 468 & 705 & \textbf{4247}\tabularnewline
\hline 
\hline 
IMMDP\cite{IMMDP} & FRCNN\cite{FasterRCNN} & 0.19 & \textbf{83.04} & 172 & 391 & \textbf{5269}\tabularnewline
\hline 
PMBM\cite{3dCNN} & Own & \textbf{0.01} & 80.39 & \textbf{121} & 1007 & 5616\tabularnewline
\hline 
extraCK\cite{extraCK} & FRCNN\cite{FasterRCNN} & 0.03 & 79.99 & 343 & 642 & 5896\tabularnewline
\hline 
MCCPD\cite{MCMOT} & FRCNN\cite{FasterRCNN} & \textbf{0.01} & 78.90 & 228 & \textbf{316} & 6713\tabularnewline
\hline 
\hline 
JCSTD\cite{JCSTD} & Regio\cite{regionlets} & 0.07 & \textbf{80.57} & 61 & \textbf{405} & \textbf{6217}\tabularnewline
\hline 
NOMT\cite{NOMT} & Regio\cite{regionlets} & 0.09 & 78.15 & \textbf{31} & 1061 & 6421\tabularnewline
\hline 
LPSVM\cite{SSVM} & Regio\cite{regionlets} & 0.02 & 77.63 & 130 & 1239 & 6393\tabularnewline
\hline 
SCEA\cite{SCEA} & Regio\cite{regionlets} & 0.06 & 75.58 & 104 & 1306 & 6989\tabularnewline
\hline 
CIWT\cite{CIWT} & Regio\cite{regionlets} & 0.28 & 75.39 & 165 & 954 & 7345\tabularnewline
\hline 
\end{tabular}
}
\end{flushright}
\caption{\textbf{Bounding box tracking results on KITTI Tracking Test (Cars).} Our method achieves the highest MOTA score of the ten best performing published methods on the KITTI tracking benchmark. Note, that the methods with FRCNN detector train the detector differently. Speed is measured in seconds per frame.}
\label{fig:table_comp_test}
\vspace{-25pt}
\end{table}

\PAR{Bounding Box Tracking: State-of-the-art comparison.}
We evaluate bounding box tracking on a KITTI MOT validation split from \cite{mots} and the official test set. The validation and test results are shown in Table \ref{fig:table_comp_val} and Table \ref{fig:table_comp_test}, respectively. MOTSFusion outperforms BeyondPixels \cite{beyondpixels}, the current best performing published method, on both sets by a significant margin, while using the same detections. It also outperforms other state-of-the-art methods that use different detections.
For the KITTI MOT definition of IDS, we reduce the IDS compared to \cite{beyondpixels} from 31 to 9 on the validation split and from 468 to 275 on the test set. When evaluating with the more challenging IDS* on the validation set, we produce only 31 IDS*, compared to 91 for BeyondPixels. MOTSFusion outperforms BeyondPixels on the MOTA metric by 0.3\% and 0.6\% on the val/test split respectively, while outperforming it by 1.6\% on the MOTA* metric on the val set. For bounding box tracking we do not perform missing detection recovery, as the checks that determine if a detection should be filled require checking against a segmentation mask of visible pixels, whereas the bounding boxes in KITTI MOT are `amodal', which means that they cover both the visible and hidden parts of objects and cannot be validated with our mask-based checks.

\PAR{Runtime.}
MOTSFusion runs in 0.44 sec. per frame, including input preprocessing, using a desktop PC with an Intel Core i7-5930K CPU and a GTX 1080 Ti GPU. Input preprocessing takes 0.20 sec. per frame, of which optical flow and depth estimation take 0.07 sec. each, and the relative ego-camera pose takes 0.06 sec. Our 2D tracklet generation takes a total of 0.07 sec. per frame to perform the mask generation, optical-flow warping and merging into tracklets. The reconstruction-based tracklet merging takes a total of 0.17 sec. of which 0.16 are used to fit the homogeneous transformations for the 3D object fusion, which is the slowest part of our tracker and could easily be sped up in the future using a GPU implementation of least-squares optimization.

In comparison to other trackers on the same hardware, TrackR-CNN \cite{mots} takes 0.50 sec. per frame, and BeyondPixels \cite{beyondpixels} takes 0.30 sec. per frame (plus 0.06 sec. for mask generation for MOTS). Thus our algorithm is able to run with a similar efficiency while performing much better, especially for long-term tracking. Our method would be faster when using LiDAR or GPS as input because depth and/or ego-motion would come directly from a sensor. As is standard for MOT, the detector runtime is not included for any of the methods.

\section{Conclusion}
We have presented a framework in which both tracking and 3D object reconstruction can be performed together and can benefit from each other, with tracking enabling reconstruction, and reconstruction enabling long-term tracking through occlusions. In evaluations, our method outperforms previous tracking methods for both bounding box and segmentation tracking. In particular, our approach is able to track objects `long-term' through complete occlusion or missed detections. We demonstrate a clear benefit of using 3D reconstruction to improve tracking, and present a tracking framework, MOTSFusion, where this can be exploited effectively and efficiently.

{%
\small
\PAR{Acknowledgments.}
This project was funded, in parts, by ERC Consolidator Grant DeeVise (ERC-2017-COG-773161).}

{
\bibliographystyle{ieee}
\bibliography{abbrev_short,egbib}
}

\end{document}